\documentclass[11pt]{article}
\pdfoutput=1
\usepackage{EMNLP2023}

\usepackage{times}
\usepackage{latexsym}

\usepackage[T1]{fontenc}
\usepackage[utf8]{inputenc}

\usepackage{microtype}

\usepackage{inconsolata}
\usepackage{graphicx}
\usepackage{tabularray}

\title{Preserving Empirical Probabilities in BERT for Small-sample Clinical Entity Recognition}

\author{  Abdul Rehman, \texttt{arehman@bournemouth.ac.uk}
\\ {\bf Jian Jun Zhang, \texttt{jzhang@bournemouth.ac.uk}}
\\ {\bf Xiaosong Yang, \texttt{xyang@bournemouth.ac.uk}}
  }

\begin{document}
\maketitle
\begin{abstract}

Named Entity Recognition (NER) encounters the challenge of unbalanced labels, where certain entity types are overrepresented while others are underrepresented in real-world datasets. This imbalance can lead to biased models that perform poorly on minority entity classes, impeding accurate and equitable entity recognition.

This paper explores the effects of unbalanced entity labels of the BERT-based pre-trained model. We analyze the different mechanisms of loss calculation and loss propagation for the task of token classification on randomized datasets. Then we propose ways to improve the token classification for the highly imbalanced task of clinical entity recognition.

\end{abstract}

\section{Introduction}

Named entity recognition (NER) is a Natural Language Processing (NLP) task of identifying and categorizing entities (such as names, events, things, and places) in a given raw text. An entity may span over just a single word or many continuous words. This process facilitates the automated analysis, search, and organization of extensive text datasets. One of the aims for NER is achieving higher precision for the entity labels with a relatively small number of training samples \cite{ding2021few}. This creates the problem of imbalance between the entities that are recognized \cite{lopez2021mining}. 

Within the clinical domain, NER holds particular significance, as it operates amidst a lexicon rich with specialized terminology necessitating accurate interpretation while imposing strict tolerances for errors. Capitalizing on the efficacy of transformer-based language models \cite{vaswani2017attention}, particularly BERT, has emerged as a dominant deep learning model for NER tasks. These models thrive on their foundational unsupervised pre-training on vast textual data, enabling them to encapsulate intricate linguistic structures that lend themselves to various language processing tasks.

The challenge of imbalanced labels considerably complicates the process of fine-tuning transformer models for Named Entity Recognition (NER). The distribution of entity classes often exhibits a significant skew, leading to disparities in the frequency of different entity types. This scenario frequently results in the underrepresentation of certain entity categories and the overrepresentation of others. Consequently, this inequality poses a hurdle to the model's capacity to generalize effectively to novel and unobserved data, particularly when it comes to the identification of essential yet infrequent entities, such as critical clinical information. In the realm of NER, it becomes imperative to address the issue of unbalanced labels, as it is pivotal to the development of models that ensure precise and equitable recognition of entities across all classes. This endeavor subsequently enhances the comprehensive efficacy and dependability of NER systems within a diverse array of practical applications.

This work makes a twofold contribution. Initially, we introduce a novel empirical bias testing methodology for BERT in token classification. We analyze the implications stemming from the application of arbitrary labels in BERT training. Furthermore, drawing on the observations and insights derived from related studies, we present a binary token labeling approach aimed at mitigating biases unsupported by empirical evidence. This enhancement seeks to augment BERT's capability to accurately discern entities characterized by a relatively low number of samples in contrast to the entities prevailing in the majority class.

\subsection{Related Works}

The proliferation of text-based data in the biomedical field, such as electronic health records, clinical documents, and pharmaceutical specifications, has led to the widespread adoption of deep learning and Natural Language Processing (NLP) methods for extracting and processing information \cite{tiwari2020assessment, li2022neural}. Additionally, studies have demonstrated that language models can partially encode clinical knowledge \cite{singhal2023large}. Contemporary generalized large-scale language models, which represent the forefront of language technology, exhibit suboptimal performance when deployed in clinical contexts, consequently undermining their reliability for clinical text analysis, as extensively noted in recent scholarly contributions \cite{hu2023zero, reese2023limitations}.
Therefore, biomedical and clinical NLP pose unique challenges, particularly the need to integrate structured domain knowledge into text representations, which is less prevalent in other domains \cite{chang2020benchmark}. To ensure the reliability of neural language modeling in the specialized medical field, models must learn directly from domain-specific terminologies rather than solely relying on general text data. As a result, significant research efforts within the medical NLP community have been devoted to integrating information from knowledge graphs into language models \cite{li2020behrt, he2022kg, naseem2022incorporating}. As a consequence, NER retains its position as the prevailing technique for clinical text analysis. However, it is noteworthy that the utilization of NER for clinical texts introduces a noteworthy challenge in the form of unbalanced accuracies. This imbalance in performance is intricately linked to the disparate distribution of data across distinct entity categories \cite{zhou2021clinical}.

The recent advancements in the field of biomedical NER, as highlighted in previous studies \cite{lee2020biobert, boudjellal2021abioner, perera2020named}, predominantly revolve around a restricted set of named entities such as diseases, chemicals, and genes. Nonetheless, it becomes imperative to broaden the scope of consideration to encompass a broader spectrum of biomedical entities. This includes entities pertinent to clinical diagnoses like diseases, symptoms, medical terms, risk factors, and vital signs, as well as epidemiological entities like infectious diseases and patient demographic information.

Certain studies in the biomedical field have explored the applicability of BERT in tasks related to biomedical NER \cite{liu2021med, lee2020biobert}, yielding remarkable levels of performance.  An analysis of BioBERT, a transformer-based model specifically refined through fine-tuning procedures within the clinical text domain, has unveiled a pivotal insight. It discerns that the principal factor contributing to erroneous inferences generated by the BioBERT model resides in its limited grasp of the domain-specific knowledge \cite{sushil2021we}.

\section{Blackbox Analysis for Empirical Uncertainty Persistence}

BERT (Bidirectional Encoder Representations from Transformers), introduced by \citealp{devlin2018bert}, represents a profound architecture grounded in self-attention mechanisms. It undergoes a pretraining phase utilizing substantial volumes of data, guided by a language modeling objective. This model yields intricate linguistic text representations that have exhibited their utility across a multitude of tasks within the domain of natural language processing. Since its inception, BERT has undergone meticulous examination and practical application across diverse domains \cite{lee2020biobert}. One of the downstream tasks for BERT is token classification, i.e., to use its contextualization capability to identify labels for words. BERT produces a 768-dimension vector for each token, processed to take into account a small amount of information about each of the other tokens in the input text. A downstream layer of a neural network can then learn to classify each token into entity categories. During fine-tuning, BERT's pre-trained weights and the final classification layer are modified to match the target task's label set. This process allows BERT to capture intricate contextual information from the input text and refine its predictions according to the token-level classification objectives, leading to robust and state-of-the-art performance across a diverse range of token-level classification tasks. In the case of NER, these labels identify types of entities that are learned from a particular annotated dataset. 

Fine-tuning BERT with unbalanced token labels is a significant consideration when adapting the model to token classification tasks where certain classes are disproportionately represented in the dataset. In scenarios where some classes occur much less frequently than others, the standard fine-tuning process can lead to biased models that perform well on majority classes but struggle with minority classes. The Unbalanced labels pose a significant challenge, as skewed distributions of entities in real-world datasets lead to accuracy issues, particularly for underrepresented classes. This imbalance affects model generalization and can hinder accurate recognition of minority entities, necessitating solutions for equitable and effective entity recognition in various applications.

\subsection{Maximum Likelihood Dilemma}

Because NER datasets can have limited labelling and significant imbalances, we contend that the conventional cross-entropy loss function, while theoretically capable of asymptotically generating the optimal token classifier based on the maximum likelihood principle, would not succeed in delivering satisfactory performance in situations characterized by imbalanced data distributions. In simpler words, if there is a training set comprising 99\% of class `O' and the remaining 1\% of class `M', there is little incentive for the optimizer to optimize in favour of class `M' if it comes at the cost of a loss for class `O'. Half of this problem is solved by using weighted cross-entropy loss. However, when a calculated loss is back-propagated, the optimizer is likely to mostly adjust the weights that were modified by the majority class because the majority class had a bigger share in the gradient during the forward pass. The layers of the neural network are oblivious to loss at the individual token level because the loss is calculated for the whole batch that contains at least one sentence.

\begin{figure}[!htb]
  \centering
  \includegraphics[width=\linewidth]{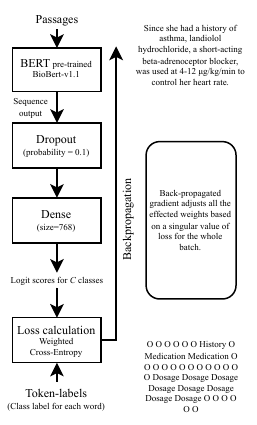}
  \caption{Framework for fine-tuning BERT for token classification using token labels for all the words. This conventional approach of learning multiple token labels at the same time can undermine the importance of sparsely occurring token labels. }
  \label{fig:BERT_backpropagation}
\end{figure}

\subsection{Exaggerated Empirical Bias in BERT for Token Classification}

\begin{figure}[!htb]
  \centering
  \includegraphics[width=\linewidth]{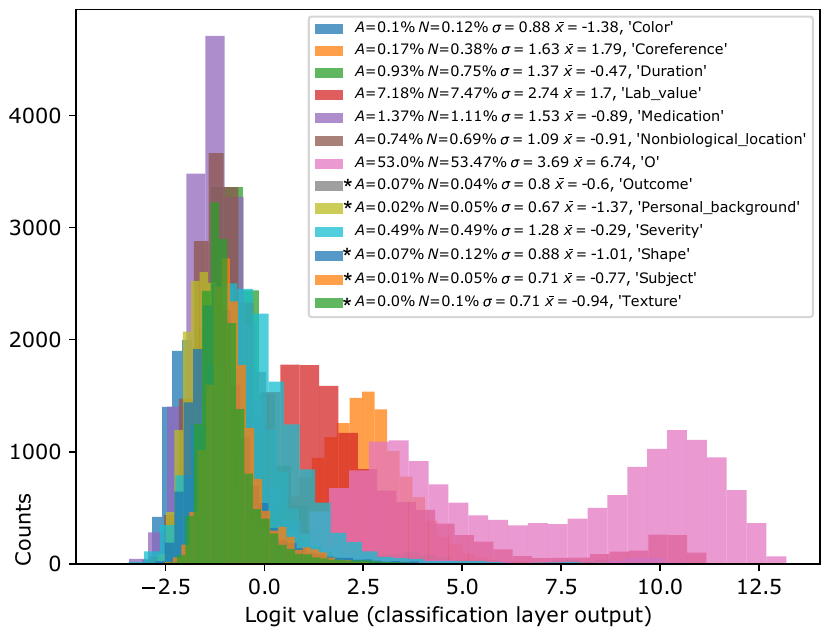}
  \caption{This histogram shows distributions of the classification layer's output logits for the clinical entity recognition task. Different colours represent entity label classes (only 13 out of 41 are shown here). $A$ is the percentage of tokens that are predicted to belong to each particular class when the logit value for that class is the highest. $N$ is the true percentage of token labels in the test set. $\sigma$ is the standard deviation and $\bar{x}$ is the mean of the corresponding logits' distributions. Small sample classes are marked by `*' as their prediction count does not match the empirically expected count. $A=0\%$ implies that none of the test samples were classified as belonging to the two smaller classes.}
  \label{fig:hist_M442_trad_BioB_W}
\end{figure}

We test the above-mentioned argument on a clinical entity dataset MACCROBAT has has 41 annotated entities ranging in occurrences from 10 to 1208 in 200 clinical documents \cite{caufield2019comprehensive, caufield2018reference, MACCROBAT}. We use 85\% of this annotated data to fine-tune BioBERT v1.1 model for 20 epochs for the token classification task as shown in Fig. \ref{fig:BERT_backpropagation}. Then we use the rest of 15\% of the documents to create a histogram of logits distributions and calculate the percentage of all the ($TP+FP$) positive predictions (represented by $A$) for each class as shown in Fig. \ref{fig:hist_M442_trad_BioB_W}.

\begin{figure}[!htb]
  \centering
  \includegraphics[width=\linewidth]{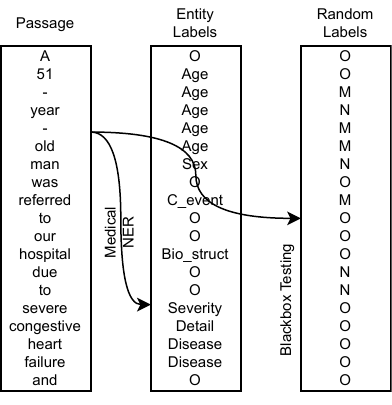}
  \caption{For the purpose of testing BERT's propensity of learning by empirical evidence, the sensible clinical entity labels are replaced with non-sensible, devoid of any meaningful pattern, randomly assigned token labels. }
  \label{fig:random_labels}
\end{figure}

\subsection{Testing BERT with Arbitrary Token Labels} \label{sec:random_labels_testing}

\begin{figure*}[!htb]
  \centering
  \includegraphics[width=\linewidth]{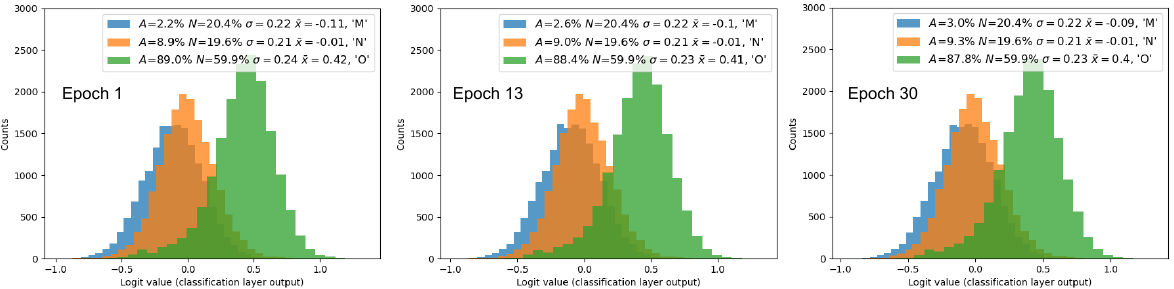}
  \caption{The distribution of predicted 3-class logits for the test set passages after training the BERT-base-cased model for 1, 13 and 30 epochs using the weighted cross-entropy loss for all token labels (ATL) in all batches. }
  \label{fig:hist_rand_trad_M442_rand1_T2_BioB_WSe7}
\end{figure*}

Assuming that there is inherent arbitrariness to language \cite{muin2021rethinking}, certain factors in languages can't be learned as hard and fast rules. We argue that if there is fuzziness in the training corpus, a good language model should retain the fuzziness if there is no significant evidence for clarity. We test this assumption on BERT by fine-tuning it on randomly generated token labels for the same clinical text corpus. We replace the original labels with randomly generated 3 classes of labels, 60\% of class `O', and 20\% each for classes `M' and `N' as shown in Fig. \ref{fig:random_labels}. Since the labels are randomly assigned, the model should not learn any significant pattern other than the unbalanced amount of labels. We train the BERT-base-cased model for 30 epochs using 85\% of the corpus and measure the logits distributions on the rest 15\% of the corpus after each epoch. We test the evidence-based uncertainty of the fine-tuned model by comparing the number of labels for each class $N$ with the number of predicted positive labels for each class represented as $A$. The difference between the two numbers will be a measure of the empirical dependency of the model since there is no other pattern to learn from other than the empirical amount of labels. Figure \ref{fig:hist_rand_trad_M442_rand1_T2_BioB_WSe7} shows the logits distributions of 3 classes for the test set. It can be observed that the model has an unaccounted bias towards the class `O' because it predicts the highest logit value for class `O' 87.8\% of the time even though empirically it should be 59.9\%.

\section{Binary Labels for Token Classification}\label{sec:equations}

When the loss is calculated by all token labels, it gets diluted into a generic loss that does not vary much from batch to batch. The problem of deeper layers being oblivious to the non-variant loss at the final layer can be solved by having only two classes creating a higher variation in loss so that it can backpropagated deeper into the model \cite{bai2020binarybert}.  We theorize that if a contextualized token classification model is fine-tuned to recognize only two labels (binary classification), then it is likely to learn each entity more precisely as compared to a model that is fine-tuned to recognize more than two labels within the same batch. We propose to use Binary Token Labels (BTL) to finetune BERT for the NER task as opposed to the conventional method of using all token labels (ATL) for each batch. The BTL method splits training batches into multiple copies but each batch of passages has binary contrastive labelling against `O' (the absence of any entity label is represented by `O') for only one true-positive label while the rest of the labels are masked. 

%$L=\left\{ 
%  \begin{array}{ c l }
%    \frac{ \exp{(x_{n, y_n})} }{\sum_{c=1}^{C}{ \exp{(x_{n, c})} } } & \quad \textrm{if } y_n  	\neq X \\
%    0                 & \quad \textrm{otherwise}
%  \end{array}
%\right.$
The weighted cross-entropy loss at the token level is calculated as
\begin{equation}
l_n= -w_{y_n}\frac{ \exp{(x_{n, y_n})} }{\sum_{c=1}^{C}{ \exp{(x_{n, c})} } }
\end{equation}
where $x_{n, c}$ is the value of the logit at the output layer for class $c$, $x_{n, y_n}$ is the logit value for the target class, $C$ is the total number classes, $w_{y_n}$ is the weight of the target class that is calculated as
\begin{equation}
w_{c} = 1 - N_c/N
\end{equation}
where $N_c$ is the number of tokens for the class $c$ and $N$ is the number of all labelled tokens in the training set.

The mean loss $L$ for the whole batch is
\begin{equation}
L = \sum_{n=1}^{B}{\frac{l_n}{\sum_{B}^{n=1}{w_{y_n}}  \cdot \{v_n \neq X \} }} 
\end{equation}

where $B$ is the number of total tokens in the batch, and $X$ is the masked label conditioned as

$v_n=\left\{ 
  \begin{array}{ c l }
    X    & y_n \notin \{O, C_b\} \\
    1  & \quad \textrm{otherwise.}
  \end{array}
\right.$

The $C_b$ is the label for batch $b$ that is left unmasked along with the true negative label $O$. The difference from the usual approach for loss calculation is that each batch contains a mixture of all target classes. Whereas in this approach, each batch only has a true positive label for only one entity class. 

\begin{figure}[!htb]
  \centering
  \includegraphics[width=\linewidth]{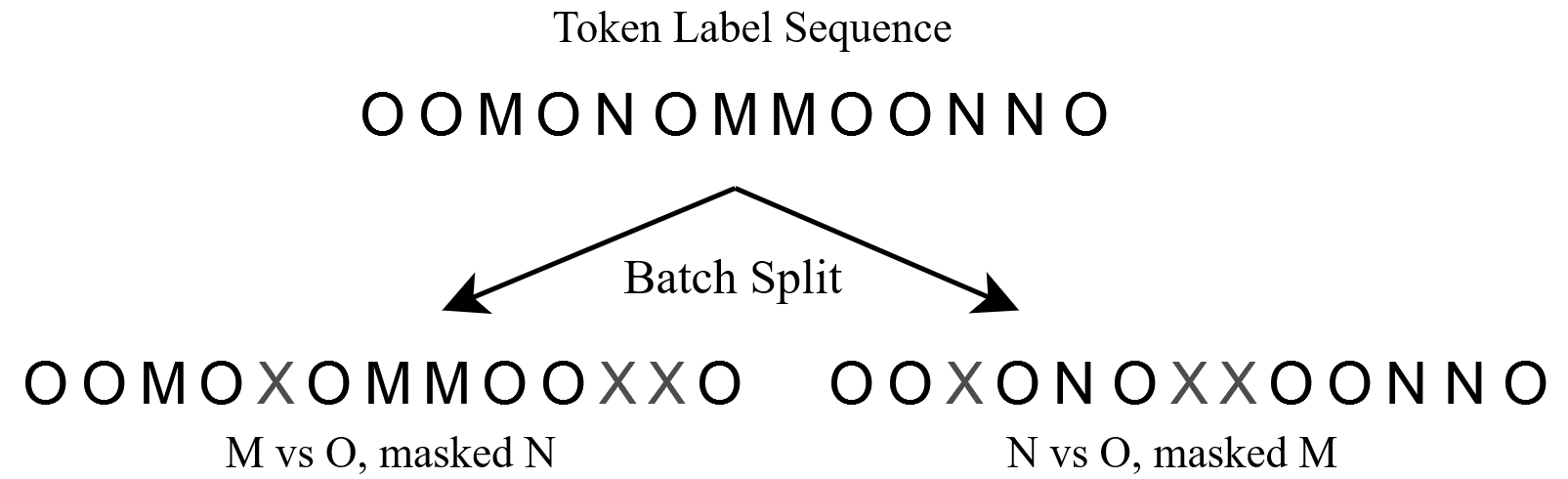}
  \caption{An example of splitting a batch to create multiple batches with only one class of true positive (either `M' or `N') per batch in contrast with true negative `O'.}
  \label{fig:Binary_tokens}
\end{figure}

\begin{figure*}[!htb]
  \centering
  \includegraphics[width=\linewidth]{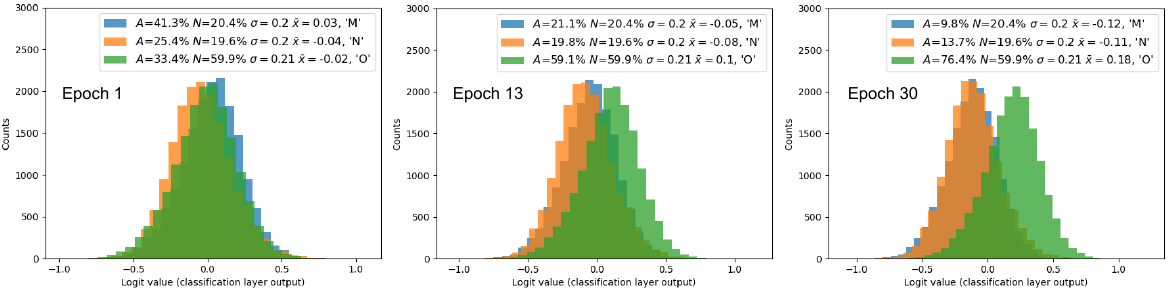}
  \caption{The distribution of predicted 3-class logits for the test set passages after training the BERT-base-cased model for 1, 13 and 30 epochs using the weighted cross-entropy loss for binary token labels (BTL) in segregated-by-class batches.}
  \label{fig:hist_rand_M442_rand2_Z2_BioB_WSe7}
\end{figure*}

\begin{figure}[!htb]
  \centering
  \includegraphics[width=\linewidth]{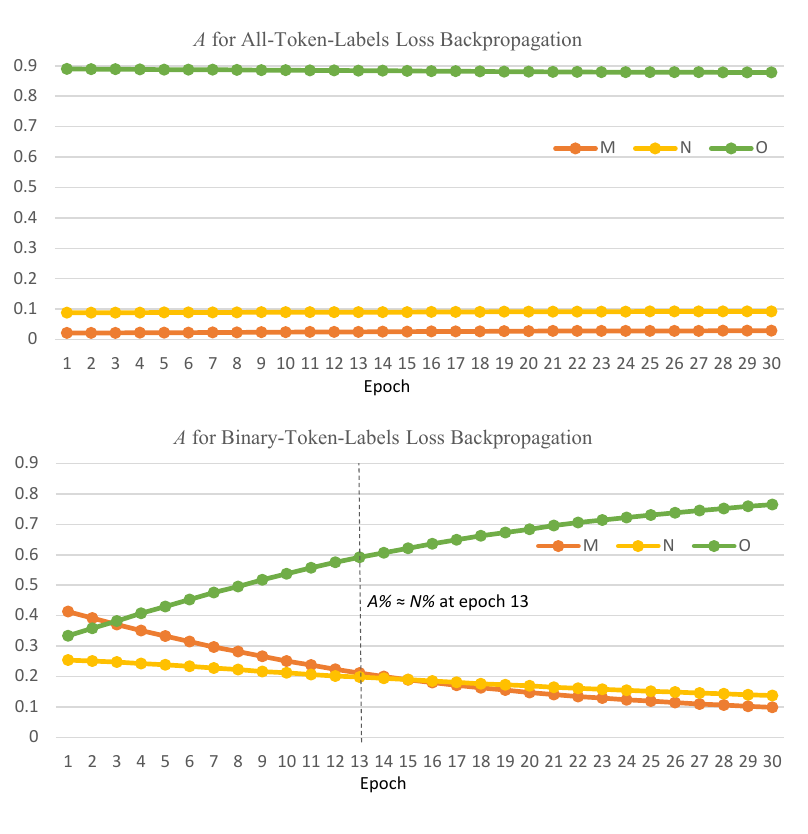}
  \caption{Plots of ratios of $A_c$, the predicted number of labels for three random entity classes at each epoch interval when fine-tuned using ATL (top) or BTL (bottom) methods. }
  \label{fig:argmax_rand_epochs}
\end{figure}

We performed the uncertainty persistence test on BERT-base-case again using the BTL approach for the randomly labelled dataset. It can be seen in Figures \ref{fig:hist_rand_M442_rand2_Z2_BioB_WSe7} and \ref{fig:argmax_rand_epochs} that using BTL the model is less sensitive to the empirical bias as it converges to the maximum-likelihood over the epochs. This makes it possible to intervene in the training process before the maximum likelihood end goal of the optimizer is reached. This approach focuses on the core task of entity presence or absence, allowing the model to learn to distinguish entities from non-entities effectively. Lastly, binary fine-tuning can yield models that are less sensitive to label noise or annotation inconsistencies for a batch.

\subsection{Experimentations}

Based on the observations made on the randomly labelled dataset in Section \ref{sec:random_labels_testing}, we tested the BTL approach on the MACCROBAT dataset with the intention that the loss caused by the majority classes won't undermine the loss caused by the small classes as it did in Figure \ref{fig:hist_M442_trad_BioB_W} when we tested the conventional token classification learning approach. The bigger documents in the MACCROBAT dataset are broken into smaller chunks to fit the maximum input tokens size of 512. This results in 200 documents being divided into 886 passages for the training set and 169 passages for the test set. Another few of the quantitative entity labels such as Volume, Mass, Height, and Weight are merged into a single category. 

To achieve a balanced performance across all entity classes we create a clinical NER method with the following measures:

\begin{itemize}
    \item Weighted cross-entropy using the class weights as calculated in Section \ref{sec:equations}.
    \item The binary token label (BTL) approach is used to create segregated batches for each entity class. We also test the conventional all-token-labels (ATL) approach where each batch has all entity labels without any masking. 
    \item The number of batches for each entity class is balanced. The minority class batches are repeated more frequently to match the number of batches of the largest class `O'.
    \item Both models (BTL and ATL) are fine-tuned using pre-trained BioBERTv1.1. The training is run for 20 epochs using an SGD optimizer with a learning rate of $5e-5$.
    \item A latent KNN classifier (17 neighbours) to predict the final entity label using the raw logits from the output layer. The KNN classifier is trained separately after finetuning.
\end{itemize}

The utilization of K-Nearest Neighbors (KNN) serves the purpose of additional independent calibration post finetuning. The outcomes for 33 entities within the MACCROBAT dataset are presented in Table \ref{tab:all_results}. Notably, employing the BTL method leads to a substantial increase in unweighted measures like unweighted accuracy and mean precision across all entities. This improvement is particularly pronounced for entities with small sample sizes.

\begin{figure}[!htb]
  \centering
  \includegraphics[width=\linewidth]{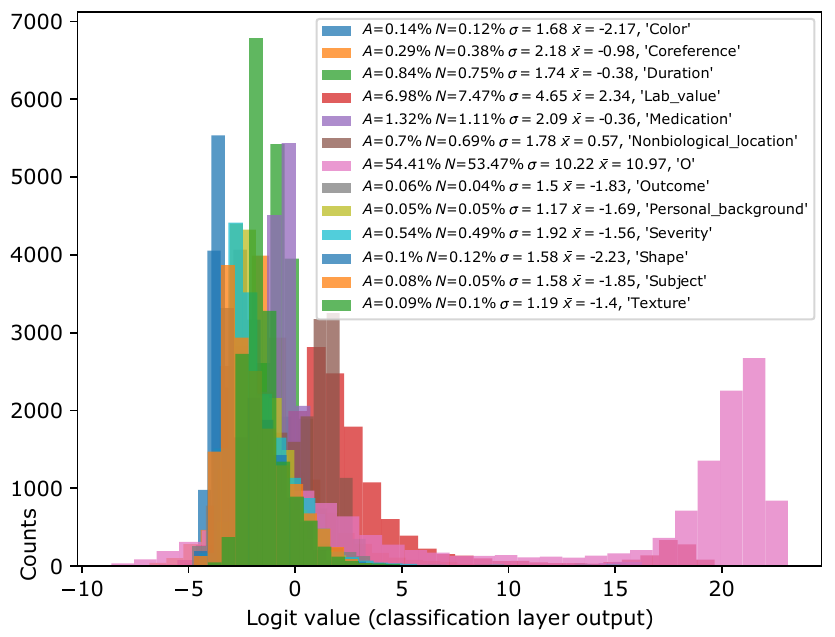}
  \caption{MACCROBAT test set's distributions of logits when BioBERT is fine-tuned using the BTL method.}
  \label{fig:hist_MBAT_Z_10e_CE_weights}
\end{figure}

Figure \ref{fig:hist_MBAT_Z_10e_CE_weights} shows the distribution of logits for the test set using the BTL approach. The number of positive predictions for small-sample entity classes are much closer to empirically expected positive predictions as compared to the distributions for ATL model (Figure \ref{fig:hist_M442_trad_BioB_W}). Another observable difference from Figure \ref{fig:hist_M442_trad_BioB_W} is the wider spread of the distributions. This is due to the increase in the entity-specific loss variation as discussed at the start of this section.

In Table \ref{tab:f1_compare}, a comparison of F1 scores between the proposed method and the baseline is provided. However, it's important to clarify that the baseline's F1-score represents the overall weighted F1 score. While examining performance metrics for select individual entities, a marked distinction from the baseline results by  \citealp{zhou2021clinical} becomes evident. The advantage of BTL is the increase in the significant increase in unweighted accuracy which is a better measure of balanced accuracy across all entities, however, it is rarely reported by other works.

\begin{table}
\centering
\tiny
\caption{Clinical NER performance metrics for the conventional fine-tuning method (ATL) and for the proposed method (BTL). }
\begin{tblr}{
  cell{1}{3} = {c=3}{c},
  cell{1}{6} = {c=3}{c},cell{40}{1} = {c=2}{},
  cell{40}{1} = {c=2}{},
  cell{40}{3} = {c=3}{c},
  cell{40}{6} = {c=3}{c},
  cell{41}{1} = {c=2}{},
  cell{41}{3} = {c=3}{c},
  cell{41}{6} = {c=3}{c},
  vline{3-4} = {1,40-41}{},
  vline{3,6} = {2-39}{},
  hline{3,37,40} = {-}{},
} \hline
                        &         & ATL-Finetuned   &      &      & BTL-Finetuned &      &      \\
Entity Label                   & N       & P      & R    & F1   & P        & R    & F1   \\

O                   & 52.13\% & 93.4          & 91.5          & 92.4          & \textbf{93.5}   & \textbf{92.4} & \textbf{92.9} \\
Diagn.\_proc.       & 8.43\%  & 88.8          & 85.1          & 86.9          & \textbf{89.2}   & \textbf{86}   & \textbf{87.6} \\
Lab\_value          & 7.16\%  & 83.3          & \textbf{88.7} & 86            & \textbf{84.7}   & 87.8          & \textbf{86.2} \\
Detail.\_desc.      & 5.58\%  & 47.5          & 53.1          & 50.3          & \textbf{52.2}   & \textbf{55.9} & \textbf{54}   \\
Bio.\_struct.       & 5.15\%  & 82.2          & 82            & 82.1          & \textbf{83.4}   & \textbf{82.5} & \textbf{82.9} \\
Sign\_symptom       & 4.36\%  & 62            & \textbf{72}   & \textbf{67}   & \textbf{63.3}   & 69.8          & 66.6          \\
Disease\_dis.       & 2.15\%  & \textbf{68.1} & 47.8          & 58            & 66.7            & \textbf{49.5} & \textbf{58.1} \\
Date                & 2.02\%  & \textbf{89.1} & 76.7          & 82.9          & 88.3            & \textbf{81.1} & \textbf{84.7} \\
Therap.\_proc.      & 1.61\%  & \textbf{75.4} & \textbf{70.7} & \textbf{73.1} & 69.6            & 67            & 68.3          \\
History             & 1.60\%  & \textbf{67.3} & 64.2          & \textbf{65.8} & 58.5            & \textbf{67.1} & 62.8          \\
Medication          & 1.53\%  & 94            & 77.2          & 85.6          & \textbf{96.2}   & \textbf{81.9} & 89.1          \\
Dosage              & 1.48\%  & 88.4          & 81.1          & 84.8          & \textbf{91.1}   & \textbf{86.4} & \textbf{88.8} \\
Age                 & 0.90\%  & 97.8          & \textbf{96.3} & 97            & 97.8            & 95.6          & \textbf{96.7} \\
Clinical\_event     & 0.76\%  & 69.1          & 78.6          & 73.9          & \textbf{73.8}   & \textbf{79.7} & \textbf{76.8} \\
Duration            & 0.73\%  & 83.1          & \textbf{79.2} & \textbf{81.1} & \textbf{84.7}   & 74.5          & 79.6          \\
Nonbio.\_loc.       & 0.70\%  & \textbf{93.9} & 87.1          & 90.5          & 92.2            & \textbf{89.8} & \textbf{91}   \\
Family\_hist.       & 0.39\%  & \textbf{79.3} & \textbf{74.8} & \textbf{77.1} & 71.6            & 72.8          & 72.2          \\
Severity            & 0.39\%  & 72            & 75.6          & 73.8          & \textbf{78}     & \textbf{79}   & \textbf{78.5} \\
Coreference         & 0.38\%  & \textbf{19}   & \textbf{41.4} & \textbf{30.2} & 14.3            & 28.1          & 21.2          \\
Distance            & 0.38\%  & 84.4          & 70.1          & 77.3          & \textbf{92.2}   & \textbf{74.7} & \textbf{83.4} \\
Quant.\_concept     & 0.31\%  & 37.2          & \textbf{76.2} & \textbf{56.7} & \textbf{46.5}   & 43.5          & 45            \\
Other\_entity       & 0.30\%  & 0             & 0             & 0             & 0               & 0             & 0             \\
Administration      & 0.24\%  & 82.8          & \textbf{75}   & \textbf{78.9} & \textbf{93.1}   & 64.3          & 78.7          \\
Area                & 0.24\%  & 71.4          & 55.6          & 63.5          & 71.4            & \textbf{100}  & \textbf{85.7} \\
Frequency           & 0.18\%  & 56            & \textbf{93.3} & 74.7          & \textbf{72}     & 85.7          & \textbf{78.9} \\
Sex                 & 0.18\%  & 100           & 92.9          & 96.4          & 100             & 92.9          & 96.4          \\
Activity            & 0.15\%  & 40            & 35.7          & 37.9          & 40              & 35.7          & 37.9          \\
Time                & 0.15\%  & 57.4          & \textbf{86.1} & 71.8          & \textbf{83.3}   & 72.6          & \textbf{78}   \\
Shape               & 0.09\%  & 25            & 41.7          & 33.3          & \textbf{55}     & \textbf{78.6} & \textbf{66.8} \\
Color               & 0.08\%  & 45            & 56.2          & 50.6          & \textbf{95}     & \textbf{82.6} & \textbf{88.8} \\
Personal\_bg.       & 0.07\%  & 75            & 75            & 75            & \textbf{100}    & \textbf{88.9} & \textbf{94.4} \\
Subject             & 0.07\%  & 11.1          & 33.3          & 22.2          & \textbf{44.4}   & \textbf{40}   & \textbf{42.2} \\
Texture             & 0.07\%  & 11.8          & 16.7          & 14.2          & \textbf{58.8}   & \textbf{66.7} & \textbf{62.7} \\
Outcome             & 0.05\%  & 83.3          & 45.5          & 64.4          & 83.3            & \textbf{55.6} & \textbf{69.4} \\

Mean                &         & 65.7          & 67            & 67            & \textbf{73.1}   & \textbf{70.8} & \textbf{70.8} \\ \hline
Overall             & 84228   & 84.5          & 84.5          & 84.5          & \textbf{85.2}   & \textbf{85.2} & \textbf{85.2} \\
Except O            & 40321   & 74.2          & 76            & 75.1          & \textbf{75.6}   & \textbf{76.6} & \textbf{76.1} \\ \hline
Weighted accuracy       &         & 74.2   &      &      & \textbf{75.57 } &               &               \\
Unweighted accuracy     &         & 64.87  &      &      & \textbf{72.66 } &               &    \\ \hline
\end{tblr}\label{tab:all_results}
\normalsize
\end{table}

\begin{table}[!htb] 
\centering
\caption{Comparison of different methods for NER task on MACCROBAT dataset.}
\begin{tabular}{lc} \hline
Method & F1-score \\  \hline
\citealp{zhou2021clinical} & 65.75 \\
BERT-ATL-Softmax & 66.7 \\
BERT-ATL-KNN & 75.1 \\
BERT-BTL-KNN & 76.1 \\  \hline
\end{tabular}\label{tab:f1_compare}
\end{table}

\section*{Limitations}
This study conducted an analysis of the empirical biases inherent in BERT's NER token classification. A novel black box testing method was introduced to assess empirical biases, independent of linguistic content, with the caveat that matching expectations through empirical evidence is valuable in domains with established parameters. Subsequently, insights gleaned from the black box testing were leveraged to enhance NER performance on a heavily imbalanced clinical dataset. While there was an overall enhancement in recognition metrics, it's worth noting that the performance improvements were not uniform across all entities. Certain entity labels, such as 'Coreference,' experienced a decrease in accuracy when utilizing the proposed approach.

\section*{Ethical Statement}
Our conducted experiments and the model framework we propose are designed to promote investigation within the clinical information extraction domain while prioritizing the prevention of privacy breaches. The data utilized in our study is publicly accessible and has been thoroughly de-identified. Although recent studies have demonstrated the challenge of reconstructing sensitive personal information from such data, a minimal potential risk exists for future models to achieve this. It's important to note that we have not made any modifications to the data's content that would enhance the probability of such an eventuality, thus ensuring the mitigation of any risks related to the leakage of private information.

% Entries for the entire Anthology, followed by custom entries
\bibliography{anthology,custom}
\bibliographystyle{acl_natbib}

\appendix

\end{document}